\documentclass{article}
\usepackage[preprint]{colm2026_conference}

\usepackage{microtype}
\usepackage{hyperref}
\usepackage{url}
\usepackage{booktabs}
\usepackage{graphicx}
\usepackage{amsmath}
\usepackage{amssymb}
\usepackage{lineno}

\definecolor{darkblue}{rgb}{0, 0, 0.5}
\hypersetup{colorlinks=true, citecolor=darkblue, linkcolor=darkblue, urlcolor=darkblue}

\title{Token Budget Saturation and Mechanistic Early-Detection of\\
Reasoning Non-Convergence in Chain-of-Thought Models}

\author{%
  Renuka Oladri \\
  Computer, Mathematical, and Natural Sciences \\
  University of Maryland \\
  College Park, MD, USA \\
  \texttt{roladri@umd.edu}
  \And
  Niveda Jawahar \\
  Computer, Mathematical, and Natural Sciences \\
  University of Maryland \\
  College Park, MD, USA \\
  \texttt{nivedaj@umd.edu}
  \And
  Abdirisak Mohamed \\
  University of Maryland \\
  SAP Labs, LLC \\
  College Park, MD, USA \\
  \texttt{amoham70@umd.edu}
}

\begin{document}

\ifcolmsubmission
\linenumbers
\fi

\maketitle

\begin{abstract}
Chain-of-thought reasoning in large language models carries a substantial inference cost, yet the relationship between allocated thinking budget and downstream accuracy remains poorly characterized.
We present two interconnected studies on DeepSeek-R1-Distill-Qwen-7B across three mathematics benchmarks (GSM8K, MATH-500, AIME 1983--2024).
In the first study, we adapt the budget-forcing mechanism of \citet{muennighoff2025s1} to constrain the model's thinking tokens and measure accuracy as a function of budget.
We find that GSM8K and MATH-500 reach 95\% of their uncapped accuracy at just 256 tokens of thinking, while AIME exhibits a pronounced bimodal pattern: 56.5\% of generations converge naturally with 96.5\% accuracy, while the remaining 43.5\% never terminate their reasoning chain even at a 10,000-token ceiling and achieve only 11.5\% accuracy.
This bimodality persists across all difficulty levels and is only weakly predicted by problem difficulty ($r^2 \approx 0.186$).
In the second study, we probe whether convergence failure is detectable from internal representations before behavioral failure manifests.
Sweeping all 28 layers at token 150, we find layer 20 yields the strongest signal (AUC $0.608 \pm 0.080$).
A nine-position checkpoint sweep (50--300 tokens) shows activation probes outperforming behavioral baselines at 8 of 9 positions (mean $\Delta$AUC $= +0.035$), a consistent directional advantage that $n=200$ is underpowered to confirm statistically at individual checkpoints.
A contamination check on post-cutoff AIME 2025 problems confirms that the convergence--correctness relationship holds on unseen data (converged accuracy 100\%, non-converged accuracy 0\%), ruling out memorization as the primary driver.
These findings suggest that reasoning models encode convergence fate in their internal state well before it becomes externally observable, motivating mechanistically-grounded early-exit strategies for inference cost reduction.
\end{abstract}

\section{Introduction}

The emergence of chain-of-thought (CoT) prompting~\citep{wei2022chain} and dedicated reasoning models such as OpenAI's o1~\citep{openai2024o1} and DeepSeek-R1~\citep{deepseek2025r1} has shifted the computational bottleneck in large language model (LLM) inference from model size to generation length.
These systems produce extended internal monologues---often thousands of tokens---before arriving at a final answer.
The prevailing assumption in the field is that longer reasoning yields better outcomes, an assumption that underpins the growing interest in test-time compute scaling~\citep{snell2024scaling}.

This assumption, however, has received limited empirical scrutiny at the level of individual generations.
While aggregate benchmarks show that reasoning models outperform their non-reasoning counterparts, it remains unclear how much of the generated thinking is actually productive.
A generation that produces 10,000 tokens of circular reasoning before timing out consumes the same compute as one that converges cleanly at 4,000 tokens---yet the two outcomes differ drastically in accuracy.

We investigate this question through two studies conducted on DeepSeek-R1-Distill-Qwen-7B, a 7-billion parameter model distilled from DeepSeek-R1 and built on the Qwen2 architecture~\citep{yang2024qwen2}.
We evaluate it across three benchmarks spanning a wide difficulty range: GSM8K~\citep{cobbe2021gsm8k} (grade school), MATH-500~\citep{hendrycks2021math} (competition-level), and a 200-problem subset of AIME 1983--2024~\citep{neubig2024aime} (olympiad-level).

In the first study (Section~\ref{sec:budget}), we adapt the budget-forcing technique of \citet{muennighoff2025s1} into a logits processor that constrains the model's thinking phase to a specified token count.
By sweeping budgets from 256 to 4,096 tokens and comparing against an uncapped baseline (10,000-token ceiling), we characterize the marginal value of additional thinking tokens across difficulty levels.
Our central finding is that accuracy saturates remarkably early on GSM8K and MATH-500, while AIME reveals a bimodal failure mode where a large fraction of generations never converge regardless of budget.

In the second study (Section~\ref{sec:detection}), we ask whether this convergence failure is predictable from the model's internal state before the failure becomes behaviorally apparent.
We instrument the model with forward hooks to capture hidden-state activations at early checkpoints in the reasoning chain, then train linear probes to predict eventual convergence.
By comparing these probes against a behavioral baseline that uses only surface-level signals (logit entropy, repetition statistics), we isolate the contribution of internal representations to early detection.

The contributions of this work are:
\begin{itemize}
\item Empirical evidence that thinking token budgets saturate at 256 tokens for GSM8K and MATH-500, challenging the assumption that more thinking is uniformly beneficial.
\item Characterization of a bimodal convergence pattern on AIME that is only partially explained by problem difficulty ($r^2 \approx 0.19$) and wastes approximately 40\% of inference compute on doomed generations.
\item Layer sweep across all 28 model layers at token 150, identifying layer 20 as the strongest signal carrier (AUC $0.608 \pm 0.080$) and revealing that convergence information peaks in upper-middle representational layers, not output-adjacent layers.
\item Nine-position checkpoint sweep (50--300 tokens) demonstrating that activation probes consistently outperform behavioral baselines at 8 of 9 positions, with the strongest early advantage at 50 tokens ($\Delta$AUC $= +0.074$), confirming that internal representations carry convergence information before surface-level signals emerge.
\item Contamination analysis on post-cutoff AIME 2025 problems showing that the convergence--correctness relationship holds on data the model cannot have memorized (converged accuracy 100\%, non-converged 0\%), validating our findings against the memorization hypothesis.
\end{itemize}

\section{Related work}

\subsection{Test-time compute and chain-of-thought reasoning}

\citet{wei2022chain} demonstrated that chain-of-thought prompting substantially improves multi-step reasoning.
\citet{snell2024scaling} showed that test-time compute scaling exhibits diminishing returns beyond certain thresholds.
The DeepSeek-R1 family~\citep{deepseek2025r1} produces long thinking chains enclosed in \texttt{<think>}$\ldots$\texttt{</think>} delimiters.
Budget forcing was introduced by \citet{muennighoff2025s1} in the s1 framework; our contribution is its systematic application across three benchmarks and extension to a token-level logits processor for fine-grained budget sweeps.

\subsection{Reasoning failure modes}

DeepSeek-R1-distilled models are known to exhibit a non-termination failure mode where the model enters a repetitive loop and never emits \texttt{</think>}, consuming the full token budget~\citep{pipis2025wait}.
\citet{pipis2025wait} show this is distillation-specific: student models loop substantially more than their teacher.
Our work adds systematic characterization of the bimodal structure of this failure and, uniquely, probes whether it is predictable from internal activations before it becomes behaviorally observable.

\subsection{Probing and mechanistic interpretability}

Linear probing of internal representations has become a standard tool for understanding what information neural networks encode at various layers~\citep{belinkov2022probing}.
In the context of language models, probes have been used to detect factual knowledge~\citep{meng2022rome}, sentiment~\citep{burns2023discovering}, and truthfulness~\citep{azaria2023internal, li2024iti}.
Our application---probing for eventual behavioral outcomes of an ongoing generation---differs from prior work in that the target variable (convergence vs.\ non-convergence) is a property of the full generation trajectory, not of the input or the current token.

\section{Budget-forcing study}
\label{sec:budget}

\subsection{Experimental setup}

\subsubsection{Model and inference configuration}

All experiments use DeepSeek-R1-Distill-Qwen-7B loaded in float16 precision on an NVIDIA RTX A5000 (24~GB VRAM).
Generation uses greedy decoding (\texttt{do\_sample=False}) to maximize reproducibility.
The model's chat template appends \texttt{<think>}\textbackslash n to the prompt, placing the model in thinking mode from the first generated token.

\subsubsection{Budget-forcing mechanism}

We implement a custom \texttt{LogitsProcessor} that monitors the generation stream for thinking-phase tokens.
The processor maintains a counter of tokens generated within the \texttt{<think>}$\ldots$\texttt{</think>} region.
When this counter reaches the specified budget $B$, the processor forces emission of the \texttt{</think>} token by setting all other logits to $-\infty$.
A critical implementation detail: the processor initializes in thinking mode (\texttt{self.thinking = True}) because the chat template already opens the thinking phase as part of the prompt, not as generated output.

\subsubsection{Benchmarks and sampling}

We evaluate on three benchmarks of increasing difficulty:
\begin{itemize}
\item \textbf{GSM8K}~\citep{cobbe2021gsm8k}: 200 problems sampled from the test split (seed=42). Grade school arithmetic.
\item \textbf{MATH-500}~\citep{hendrycks2021math}: 200 problems sampled from the test split (seed=42). Competition-level mathematics.
\item \textbf{AIME 1983--2024}~\citep{neubig2024aime}: 200 problems sampled from the full 933-problem dataset (seed=42). American Invitational Mathematics Examination, olympiad-level.
\end{itemize}

For each benchmark, we run generation at budgets $B \in \{256, 512, 1024, 2048, 4096\}$ tokens plus an ``uncapped'' condition. In all conditions, a practical ceiling of 10,000 total tokens is imposed; we refer to the unbudgeted condition as uncapped with this caveat, revisited in Section~\ref{sec:limitations}.
Answer extraction uses cascading regex patterns: \texttt{\textbackslash boxed\{\}} first, then ``the answer is'' patterns, then the final numeric token after \texttt{</think>}.

\subsubsection{Metrics}

We define the \emph{minimum sufficient budget} $B^*$ as the smallest budget at which the model achieves $\geq$95\% of its uncapped accuracy, sustained at all higher budgets.
We compute 95\% confidence intervals for $B^*$ via bootstrap resampling (1,000 iterations).
For binary proportions (accuracy, convergence rate), we report Wilson score intervals.

\subsection{Results}

\subsubsection{Budget saturation on GSM8K and MATH-500}

Table~\ref{tab:saturation} presents the main budget-forcing results.
Both GSM8K and MATH-500 reach their $B^*$ at the smallest tested budget of 256 tokens.
Accuracy is essentially flat across all budget levels, including uncapped.

\begin{table}[t]
\centering
\begin{tabular}{lccc}
\toprule
\textbf{Benchmark} & $\mathbf{B^*}$ & \textbf{95\% CI} & \textbf{Uncapped Acc.} \\
\midrule
GSM8K    & 256  & {[256, 512]}  & 0.740 \\
MATH-500 & 256  & {[256, 4096]} & 0.625 \\
AIME     & ---  & ---           & 0.595 \\
\bottomrule
\end{tabular}
\caption{Minimum sufficient budget ($B^*$) and uncapped accuracy.}
\label{tab:saturation}
\end{table}

This result challenges the intuition that reasoning models benefit from extended thinking on routine mathematical tasks.
The marginal compute spent beyond 256 tokens yields no measurable accuracy improvement on either benchmark, suggesting that the model's answer is effectively determined within the first 256 tokens of thinking.

AIME does not yield a finite $B^*$ because accuracy never stabilizes---the bimodal convergence pattern ensures that non-converging generations always drag down the aggregate regardless of budget.

\subsubsection{Bimodal convergence on AIME}

The AIME results reveal a striking bimodal pattern not present in the other benchmarks.
Generations split cleanly into two groups: those that produce a \texttt{</think>} token and terminate (converged), and those that never emit \texttt{</think>} within the 10,000-token ceiling (non-converged).
Table~\ref{tab:bimodal} shows the aggregate statistics.

\begin{table}[t]
\centering
\begin{tabular}{lcccc}
\toprule
\textbf{Group} & $n$ & \textbf{Share} & \textbf{Accuracy} & \textbf{Avg Tokens} \\
\midrule
Converged     & 113 & 56.5\% & 96.5\% & $\sim$4,100 \\
Non-converged &  87 & 43.5\% & 11.5\% & 10,000 \\
\bottomrule
\end{tabular}
\caption{AIME convergence analysis (uncapped, $n=200$).}
\label{tab:bimodal}
\end{table}

The accuracy gap between the two groups is enormous: convergence is nearly synonymous with correctness (96.5\%), while non-convergence nearly guarantees failure (11.5\%).
Non-converged generations consume the full token budget while contributing almost nothing to accuracy.

\subsubsection{Convergence is not explained by difficulty}

A natural hypothesis is that non-convergence simply reflects problem difficulty---the model gets stuck on hard problems.
Point-biserial correlation between convergence status and problem number yields $r = 0.431$ ($p < 0.0001$), confirming a moderate relationship, but $r^2 \approx 0.186$ means that problem order explains less than 19\% of convergence variance.

Table~\ref{tab:difficulty} breaks down accuracy by difficulty tercile, showing that the convergence--accuracy gap persists even within each tercile.

\begin{table}[t]
\centering
\begin{tabular}{lcc}
\toprule
\textbf{Difficulty Tercile} & \textbf{Converged Acc.} & \textbf{Non-conv.\ Acc.} \\
\midrule
Easy (Problems 1--5)    & 95.5\%          & 18.2\% \\
Medium (Problems 6--10) & 95.1\%          & 11.1\% \\
Hard (Problems 11--15)  & 100\% ($n=18$)  & 0\% ($n=41$) \\
\bottomrule
\end{tabular}
\caption{Accuracy by convergence status and difficulty tercile.}
\label{tab:difficulty}
\end{table}

The hard tercile is particularly striking: every converged generation on a hard problem is correct, while non-converged generations achieve nothing.
This suggests convergence and correctness are nearly co-extensive on difficult problems.

\subsubsection{Practical implications}

Non-converged generations consume approximately 10,000 tokens each while contributing almost zero correctness signal.
If an oracle could identify non-converging generations at generation start and abort them, approximately 43.5\% of AIME inference compute could be saved.
The early-detection study (Section~\ref{sec:detection}) investigates how closely a probe-based system can approximate this oracle.

\section{Early detection study}
\label{sec:detection}

\subsection{Experimental setup}

\subsubsection{Instrumented generation}

We regenerate all 200 AIME problems with the same configuration (greedy decoding, seed=42, 10,000-token ceiling), adding two forward hooks:
\begin{itemize}
\item A \textbf{hidden-state hook} on all 28 layers at token position 150, capturing activation vectors across the full depth of the model.
\item A \textbf{logit hook} on the \texttt{lm\_head} layer, computing token-level entropy on the fly at each of nine checkpoint positions (50, 75, 100, 125, 150, 175, 200, 250, 300 tokens).
\end{itemize}

A critical implementation concern is the distinction between prefill and autoregressive steps in \texttt{transformers}: the hook fires for every forward pass, including the full prompt prefill.
We track the generation step count to ensure hooks capture only the autoregressive steps at the specified positions, not prefill activations.

Each sample's convergence label and activation are captured from the \textbf{same} generation run.

\subsubsection{Fresh run validation}

Table~\ref{tab:sanity} compares the fresh instrumented run against the original budget-forcing results to verify consistency.

\begin{table}[t]
\centering
\begin{tabular}{lcc}
\toprule
\textbf{Metric} & \textbf{Fresh Run} & \textbf{Original} \\
\midrule
Convergence rate      & 62.0\% & 56.5\% \\
Converged accuracy    & 90.3\% & 96.5\% \\
Non-conv.\ accuracy   &  6.6\% & 11.5\% \\
Avg tokens (conv.)    & 4,889  & $\sim$4,100 \\
Avg tokens (non-conv.)& 10,000 & 10,000 \\
\bottomrule
\end{tabular}
\caption{Sanity check: fresh instrumented run vs.\ original budget-forcing run.}
\label{tab:sanity}
\end{table}

The qualitative pattern is preserved: strong bimodal split, high convergence accuracy, near-zero non-convergence accuracy.
The minor quantitative differences (62.0\% vs.\ 56.5\% convergence rate) are within expected sampling variability for $n=200$.

\subsubsection{Probe design}

\paragraph{Activation probe.}
A logistic regression classifier ($C=1.0$, L-BFGS solver, 2,000 max iterations) trained on the hidden-state activation vector at a given layer and checkpoint position.
The input dimensionality equals the model's hidden size (3,584 for DeepSeek-R1-Distill-Qwen-7B).

\paragraph{Behavioral baseline.}
A logistic regression classifier with identical hyperparameters, trained on seven surface-level features observable without model internals:
\begin{itemize}
\item Entropy statistics: mean, maximum, standard deviation, and linear trend (slope of entropy over generated tokens).
\item N-gram repetition rates: fraction of repeated bigrams and trigrams in the generated token sequence.
\item Raw token count at the checkpoint.
\end{itemize}

These features capture the information available to a system monitoring the model's output stream without access to internal states.

\paragraph{Combined probe.}
A logistic regression classifier trained on the concatenation of the activation vector and the seven behavioral features.

\subsubsection{Evaluation protocol}

Given the modest sample size ($n=200$, with approximately 124 converged and 76 non-converged samples), we use 5-fold stratified cross-validation with AUC-ROC as the primary metric.
We apply the correction $\max(\text{AUC}, 1 - \text{AUC})$ to account for label-flip ambiguity in linear probes.
Per-checkpoint significance is assessed via a paired permutation test ($B=10{,}000$ permutations) on fold-level AUC differences. Because the nine checkpoints are not independent (they derive from the same 200 generations), we additionally run a sweep-level permutation test that flips activation and behavioral labels identically across all checkpoints, computing the mean $\Delta$AUC under $B=100{,}000$ permutations.

\subsection{Results}

\subsubsection{Checkpoint position sweep}

\begin{figure}[t]
\centering
\includegraphics[width=\linewidth]{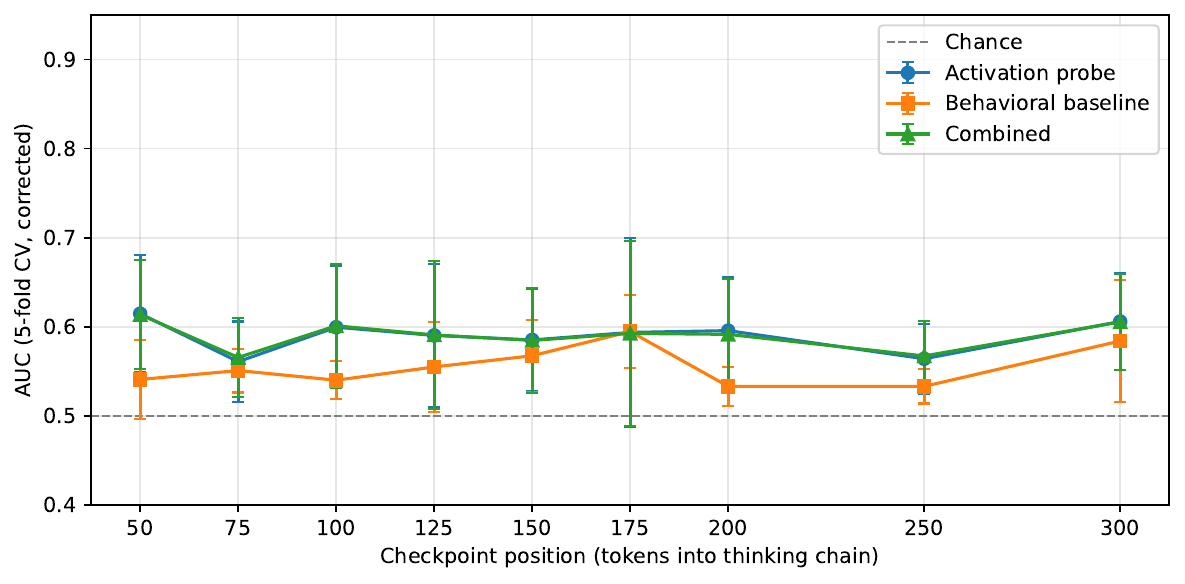}
\caption{Activation probe, behavioral baseline, and combined probe AUC vs.\ checkpoint position (5-fold CV, layer 16, $n=200$). Error bars show $\pm$1 standard deviation across folds. The activation probe (blue) consistently leads the behavioral baseline (orange) across all positions; the combined probe (green) tracks the activation probe closely, indicating that behavioral features add little once activations are included.}
\label{fig:checkpoint}
\end{figure}

Figure~\ref{fig:checkpoint} shows probe AUC across nine checkpoint positions (50--300 tokens), captured in a single instrumented generation run with layer 16 hooked throughout. Layer 16 was chosen as a mid-depth default before the layer sweep was run; the layer sweep subsequently identified layer 20 as the strongest signal carrier (Section~\ref{sec:layersweep}), so the checkpoint sweep likely underestimates the peak achievable AUC by a small margin.

\subsubsection{Consistent directional advantage across checkpoints}

Across all nine checkpoint positions, activation probes outperform behavioral baselines at 8 of 9 positions (all except 175 tokens, where they are essentially tied at $\Delta$AUC $= -0.001$).
The mean advantage across all positions is $\Delta$AUC $= +0.035$.
The strongest early advantage is at 50 tokens (activation AUC $= 0.615 \pm 0.066$ vs.\ behavioral $0.541 \pm 0.044$, $\Delta = +0.074$).

First, the behavioral baseline is weak at every position, hovering near chance throughout the sweep (AUC 0.533--0.595), with its lowest values in the 50--100 token window.
At this early stage, surface-level statistics carry little information about eventual convergence: logit entropy at token 50 reflects the model's uncertainty about the next token, not about whether the overall reasoning trajectory will terminate, and repetition has not yet accumulated enough to be discriminative.

Second, activation probe performance is consistent across checkpoints (AUC 0.561--0.615) and folds, with its largest advantage over the behavioral baseline precisely where the baseline is weakest ($\Delta = +0.074$ at 50 tokens, $\Delta = +0.059$ at 100 tokens).
A paired permutation test on the mean $\Delta$AUC across all nine checkpoints, constructed to preserve the dependence between checkpoints (the same 200 generations underlie every position), yields $p = 0.063$, consistent with a modest signal that this sample size cannot confirm at conventional thresholds.

While no individual checkpoint reaches statistical significance given our sample size, the sweep-level permutation test ($p = 0.063$) and the consistent directionality (8 of 9 positions positive) together constitute evidence for a real, if modest, signal in the activation space.

\subsubsection{Signal dynamics across checkpoints}

The sweep does not reveal a monotone trend in either probe: the behavioral baseline remains weak across the full 50--300 token range (AUC 0.533--0.595), and the activation probe's AUC fluctuates within a narrow band (0.561--0.615) rather than growing with additional context.
The practical implication is favorable for early exit: since the activation signal is already at full strength by 50 tokens---the position where the behavioral baseline is closest to chance---nothing is gained by waiting.
A probe deployed at 50 tokens could in principle flag non-converging generations roughly 4,000--9,000 tokens before behavioral failure becomes unambiguous (non-emission of \texttt{</think>} near the ceiling), representing the maximum possible compute saving, subject to the discrimination limits discussed in Section~\ref{sec:limitations}.

\subsubsection{Layer sweep results}
\label{sec:layersweep}

\begin{figure}[t]
\centering
\includegraphics[width=\linewidth]{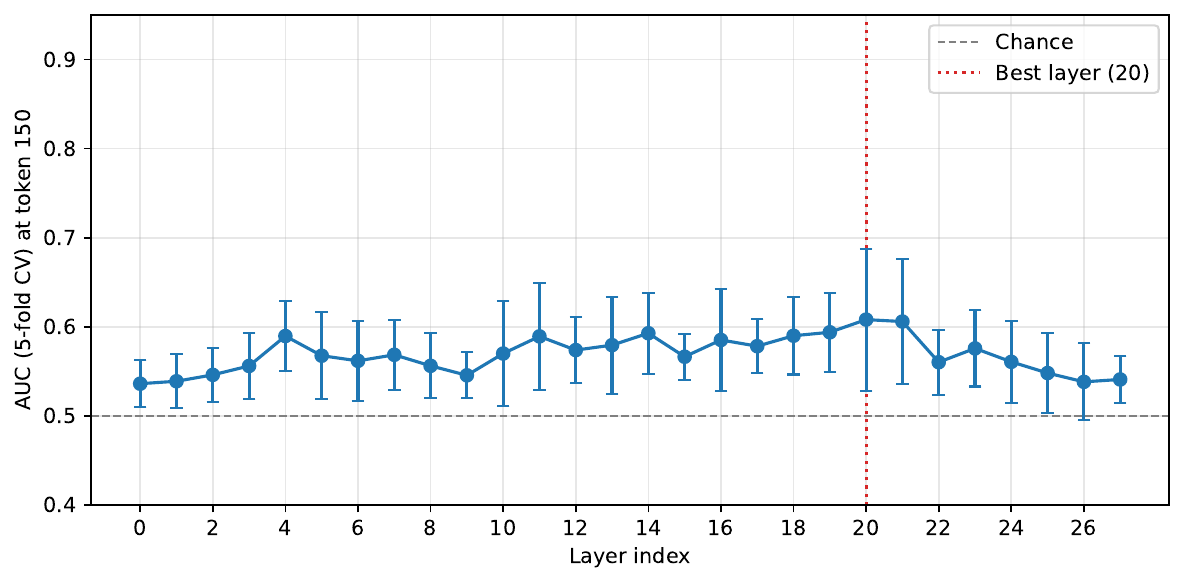}
\caption{Probe AUC across all 28 layers at token 150 (5-fold CV). The signal peaks at layer 20 (AUC $0.608 \pm 0.080$) and drops toward chance in early ($\leq$3) and final ($\geq$25) layers, indicating that convergence information concentrates in upper-middle representational layers.}
\label{fig:layer}
\end{figure}

To identify which layer carries the strongest convergence signal, we probe all 28 layers at token 150 independently (Figure~\ref{fig:layer}).
The AUC rises from near-chance in early layers (layer 0: AUC $0.536$), peaks in the upper-middle layers (layer 20: AUC $0.608 \pm 0.080$; layer 21: AUC $0.606 \pm 0.070$), and drops back toward chance in the final layers (layer 27: AUC $0.541$).
This pattern suggests that convergence information is encoded in the upper-middle representational layers of the transformer, not in the output-adjacent layers most directly tied to token prediction.

\subsubsection{Contamination check on AIME 2025}

A concern with AIME 1983--2024 results is that the model may have memorized historical problems during training (training cutoff: July 2024).
To test this, we run uncapped inference on all 30 AIME 2025 problems (released February 2025, strictly after the training cutoff) using the same model and greedy decoding.
Table~\ref{tab:contamination} compares the results.

\begin{table}[t]
\centering
\begin{tabular}{lcc}
\toprule
\textbf{Metric} & \textbf{AIME 1983--2024} & \textbf{AIME 2025} \\
\midrule
Convergence rate       & 62.0\% & 36.7\% \\
Overall accuracy       & 58.5\% & 36.7\% \\
Converged accuracy     & 90.3\% & 100.0\% (11/11) \\
Non-converged accuracy &  6.6\% & 0.0\% (0/19) \\
\bottomrule
\end{tabular}
\caption{Contamination check: AIME 2025 vs.\ AIME 1983--2024.}
\label{tab:contamination}
\end{table}

The lower overall accuracy on AIME 2025 (36.7\% vs.\ 58.5\%) is attributable to a lower convergence rate (36.7\% vs.\ 62.0\%), not a collapse in per-problem reasoning quality.
When the model converges on a 2025 problem, it achieves 100\% accuracy---higher than the 90.3\% on historical problems.
When it does not converge, accuracy is 0\%, compared to 6.6\% on historical data.
The 6.6\% non-convergence accuracy on historical problems likely reflects partial memorization of answer values; this effect disappears entirely on post-cutoff data, as expected.
The reduced convergence rate on 2025 problems is consistent with their being harder or more out-of-distribution, not with the model relying on memorized answers.

\section{Discussion}

\subsection{Rethinking token budgets}

The budget saturation results on GSM8K and MATH-500 carry implications for deployment.
Current reasoning models generate thousands of thinking tokens by default, but our results suggest that for routine mathematical reasoning, 256 tokens of structured thinking suffice.
The marginal compute spent on additional tokens yields no measurable benefit and represents a pure efficiency loss.

The AIME results complicate this picture.
The absence of a finite $B^*$ does not mean that more budget helps; rather, it reflects the bimodal failure mode where additional tokens do not rescue non-converging generations.
The problem is not insufficient budget but a qualitatively different failure mode that more compute cannot resolve.

\subsection{Nature of the non-convergence signal}

The early-detection results raise questions about what the linear probe is detecting in the hidden state at 150 tokens.
Several hypotheses are plausible:
\begin{itemize}
\item \textbf{Problem representation quality}: The model may form a more or less coherent representation of the problem during the first 150 tokens of reasoning, and this representation quality is predictive of eventual convergence. A poorly-formed representation leads to circular reasoning.
\item \textbf{Strategy selection}: The model may commit to a solution strategy early. Productive strategies lead to convergence; unproductive ones lead to loops. The hidden state at 150 tokens may encode which strategy class was selected.
\item \textbf{Confidence calibration}: The hidden state may encode the model's implicit assessment of whether its current reasoning trajectory is productive, even before this assessment manifests in output token statistics.
\end{itemize}

Distinguishing these hypotheses would require more targeted interpretability techniques (e.g., activation patching~\citep{meng2022rome}) and is beyond the scope of this work.

\subsection{Toward practical early-exit systems}

An operationally useful early-exit system would require substantially higher discrimination than the best observed AUC ($\sim$0.61), but the current results establish a foundation.
Immediate directions include nonlinear probes (MLP or attention-based classifiers) as our linear probe is a lower bound on extractable information; combining activations across multiple layers rather than a single layer, given that layer 20 is the strongest but not the only signal carrier; and temporal integration across a sliding window of checkpoints rather than a single position.
Larger evaluation sets ($n \gg 200$) would also tighten estimates and enable more expressive architectures.

\section{Limitations}
\label{sec:limitations}

\textbf{Single model.} All experiments use DeepSeek-R1-Distill-Qwen-7B. The bimodal convergence pattern, budget saturation behavior, and early-detection signal may not generalize to other model families, scales, or training procedures.

\textbf{Practical token ceiling.} The ``uncapped'' condition imposes a 10,000-token ceiling. While non-converged generations show no signs of approaching convergence at this limit, we cannot rule out that some would converge given substantially more tokens.

\textbf{Sample size and statistical power.} With $n=200$ and cross-validation folds of $\sim$40 samples, power is limited; cell sizes reach $n=18$ (converged-hard) and $n=4$ (some bootstrap cells). The probe also demonstrates that predictive information is \emph{present} in the hidden state, not that it \emph{causally determines} convergence; establishing causality requires intervention experiments such as activation patching.

\textbf{Greedy decoding.} All generations use greedy decoding (\texttt{do\_sample=False}) to maximize reproducibility. DeepSeek's own usage recommendations for the R1 series advise against greedy decoding, noting it can cause repetitive generation, and suggest a temperature around 0.6. This raises the question of whether the 43.5\% non-convergence rate is an intrinsic model property or a decoding artifact. We note that the bimodal structure we document---clean \texttt{</think>} termination versus unbounded repetition without any partial termination---is consistent with \citet{pipis2025wait}, who document the same loop failure mode in R1-distill under standard inference settings. Nonetheless, a direct comparison run at temperature 0.6 across multiple seeds would strengthen the causal claim and is left for future work.

\textbf{Potential memorization on historical data.} Our contamination check on AIME 2025 shows that 6.6\% of non-converged historical generations are correct, likely due to partial memorization of answer values. This effect is absent on post-cutoff data, suggesting our historical baseline slightly overestimates the model's true mathematical capability on novel problems.

\textbf{Difficulty proxy.} AIME problem number serves as an imperfect difficulty proxy. Problems are not uniformly ordered by difficulty within each exam, and difficulty varies across exam years.

\section{Conclusion}

We have presented evidence that chain-of-thought reasoning models exhibit two distinct inefficiency patterns: budget saturation, where accuracy plateaus at a small fraction of the model's maximum thinking length, and bimodal convergence failure, where a substantial fraction of generations enter non-terminating loops that consume the full token budget without contributing to accuracy.

More strikingly, we have shown that the convergence outcome is partially encoded in the model's internal representations well before it manifests behaviorally.
A linear probe on hidden-state activations shows a consistent directional advantage over behavioral baselines at 8 of 9 tested checkpoints, with signal detectable as early as 50 tokens---before any surface-level indicators of failure appear.
The strongest signal is found at layer 20 (of 28), suggesting convergence information is encoded in upper-middle representational layers rather than output-adjacent ones.

These findings point toward a class of inference-time optimization strategies that are grounded in the model's own internal representations rather than in behavioral heuristics.
While the current probe accuracy is insufficient for production deployment, the existence of a consistent, if modest, early signal establishes the feasibility of mechanistic early-exit approaches and motivates further investigation with more expressive probes, larger datasets, and causal validation.

\section*{Data and code availability}

All code, instrumented generation pipelines, and results will be made available upon acceptance of the paper.

\bibliography{paper}
\bibliographystyle{colm2026_conference}

\end{document}